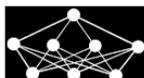

# SITUATION MODEL OF THE TRANSPORT, TRANSPORT EMISSIONS AND METEOROLOGICAL CONDITIONS

V. Beneš*, M. Svitek*, A. Michalíková †$, M. Melicherčík †

**Abstract:** Air pollution in cities and the possibilities of reducing this pollution represents one of the most important factors that today's society has to deal with. This paper focuses on a systemic approach to traffic emissions with their relation to meteorological conditions, analyzing the effect of weather on the quantity and dispersion of traffic emissions in a city. Using fuzzy inference systems (FIS) the model for prediction of changes in emissions depending on various conditions is developed. The proposed model is based on traffic, meteorology and emission data measured in Prague, Czech Republic. The main objective of the work is to provide insight into how urban planners and policymakers can plan and manage urban transport more effectively with environmental protection in mind.

## 1. Introduction

Modelling of emission effects in cities is being studied in number of research works. In this section of the paper, we present brief overview of modern literature related to this topic.

In this paper we are focused on a Smart City 4.0 framework based on the principles of Industry 4.0, that integrates computing technologies to enhance city infrastructure, data collection, and utilization, aiming to improve city operations [1].

In this case it also follows the definition of the Intelligent Transport Systems (ITS), that are designed through interconnected system components and functions, utilizing mathematical tools for optimizing performance, telecommunications, and economic efficiency, with a strong emphasis on cost-benefit analysis to assess ITS effectiveness [2].

* Viktor Beneš; Miroslav Svitek: Czech Technical University in Prague, Faculty of Transportation Sciences, Konviktská 20, CZ-11000 Praha 1, Czech Republic, E-mail: `benesvi1@cvut.cz`; `miroslav.svitek@cvut.cz`

† Alžbeta Michalíková – Corresponding author; Miroslav Melicherčík: Matej Bel University, Faculty of Natural Sciences, Tajovského 40, SK-974 01 Banská Bystrica, Slovakia, E-mail: `alzbeta.michalikova@umb.sk`; `miroslav.melichercik@umb.sk`

$ Alžbeta Michalíková – Corresponding author: Mathematical Institute, Slovak Academy of Sciences Dumbierska 1, SK-974 01 Banská Bystrica, Slovakia, E-mail: `alzbeta.michalikova@umb.sk`

The idea of this research work is continuation of the study [3] which focuses on the Knowledge Graph in the context of Smart Cities. This work is focused on the specification of one chosen city (Prague, Czech Republic). Authors selected several heterogeneous areas of a city with the objective of understanding relationships between them. One of the continuations of this approach was the research of Knowledge graphs for transport emissions concerning meteorological conditions, mentioned in [4].

Based on these results the need for multiple data sources and categorization of the data arises. This paper expands dataset used in [4] which concerns:

- the direction and speed of the wind,
- categorization of meteorological condition data such as temperature and rainfall,
- the level of service (LoS) of traffic flow.

This paper is focused on automating data processing for a defined Prague location to design a system that can be potentially applicable globally. This paper integrates data from public companies that combine traffic flow, meteorological situation, and emission information. Through data preprocessing and categorization, we defined an analytical "Situation model" reflecting driver and emission behavior under various meteorological conditions. This model suggests real-time urban actions, like traffic management based on weather forecasts, aiming for efficient city planning and improved environmental conditions.

Main objective of this paper is to use variables describing traffic situation in Prague, which are part of the mentioned dataset and model a Level of Service degree using specific fuzzy approach – Takagi-Sugeno fuzzy inference system. The proposed system focuses on the specific city (Prague, Czech Republic) and the behavior of commuters on the street with the most traffic flow in the city. Significant benefit of this approach is the automatic data processing that can be done for future research in this field to set up the system for another location.

The body of this study is structured as follows: Section 2 describes system approach to the selected data. Section 3 presents fuzzy approach to model Level of Service in the city based on traffic data. Sections 4 and 5 contain the discussion and a conclusion to the work.

## 2. System approach

The automatization of each process is the basis for design of complex system. Our objective is to design and implement system which is based on the automatic data processing for selected location in Prague. If working correctly, such system can be used in other locations and, therefore, create global city model.

Figure 1 introduces the general scheme of the system. As the input, the system introduces public companies TSK (Technical Manager of Roads), OICT (Operator ICT), CHMU (Czech Hydrometeorological Institute), and IPR (Institute of planning and Development). Each of these public companies brings up part of the needed data. TSK and OICT have the traffic flow data, OICT and CHMU have the meteorological data, and IPR and CHMU have the emission data.

The data must be preprocessed first. Data preprocessing helps in modification of data before further analysis. It is an important step in the data-gathering process.

In [4] authors used Knowledge Graphs to model the situation of the Prague crossroads. Such knowledge is necessary to categorize input data. There should be defined rules based on the data for the categorization of each data sample.

Based on the data and its categorization the analytical model called the Situation model is developed. It is a model of the situation based on the behavior of drivers and emissions based on meteorological conditions. The results of this model can propose actions in real city situations.

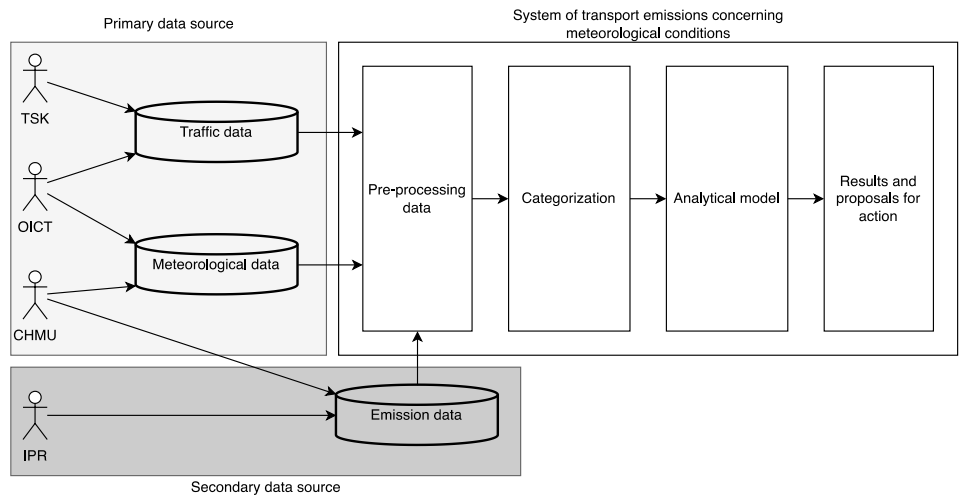

**Fig. 1:** *Scheme of the system for transport, emissions, and weather conditions.*

The detailed description of the system is in Figure 2. The scheme presents the system and its future development using live or estimated data and can estimate the action in real city situations. There are two possible actions however it is not the complete list but the possibilities.

Gathering data is focused on historical data for as long time series as possible. There is a need to understand the behavior of each data attribute for each location. The categorization is managed for the traffic flow and the meteorological conditions. For the emission data, it is using still absolute values. The traffic categorization is based on the Level of service values (1 – as the best value and the 5 for the congestion). The meteorological conditions are categorized mainly for the temperature and rainfall based on the geographical location. The categorization is set as for example winter day, spring day and so for the temperature and day without rain or day with light rain for the rainfall.

The categorization helps to use the regression or other statistical approach to the data. The analysis for each location creates a situation model that can be extrapolated for the whole monitored area.

The situation model can be then used as the method for predicting the future situation. For example – the driver during the morning must decide the usage of the car during the day. The city can manage the traffic situation based on the weather forecast – change the price of parking – motivate people to park on the P+R parking and demotivate people to park in the city center.

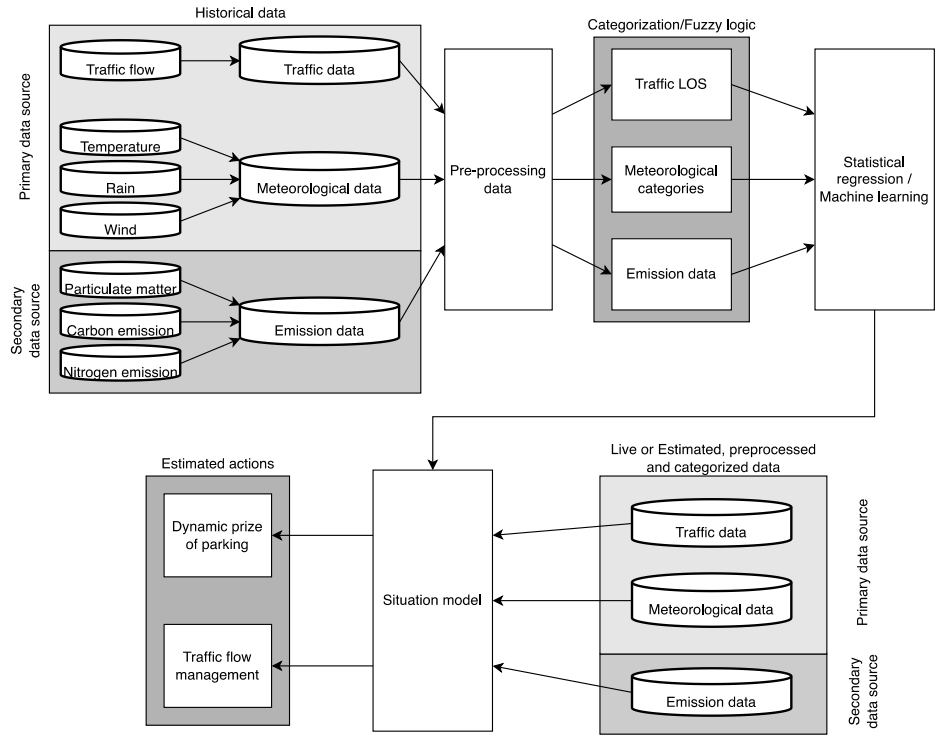

**Fig. 2:** *Scheme of the detailed system for transport, emissions, and weather conditions using live data.*

## 3. Fuzzy inference systems

Fuzzy inference systems represent the models that are able to work with human knowledge. They are constructed in such a way that they can capture knowledge expressed by ordinary speech, transform it into IF-THEN rules and then model non-linear systems of any complexity [5]. Therefore, they are highly suitable to model the solution to the presented problem. On the other hand, fuzzy systems use simple operations, for example max, min, weighted average, to get results from the IF-THEN rules. We aim to develop a fuzzy inference system for each type of mentioned data – traffic, meteorological, emission – separately and then combine these outputs into the final result. For example, similar approach was used in [6] where the outputs of more ANFISs were combined and used to predict and analyze the mutual relationship between the selected variables which affect carbon dioxide emissions. Similarly, in [7] the so called Hierarchical fuzzy inference system was used to model urban air pollution.

In this part, we present the way of the traffic data processing. This part of dataset is the simplest one since it contains fewer inputs and therefore, it is the most fitting to represent the mentioned approach. These data are usually described by a value of "LoS" (Level of Service), with parameters "speed of traffic flow" (measured in kilometers per hour) and "traffic flow" (measured in vehicles per hour, see Figure 3). LoS is also affected by traffic flow on cross streets and left turn signal phases which are not considered in this study.

The Czech standards [8] help us to categorize the traffic flow data into the six categories of LoS. The definition of each category is given in the form of word description of each level:

- LoS 1 – The traffic flow is free.
- LoS 2 – Traffic flow is almost continuous.
- LoS 3 – The traffic situation is stable.
- LoS 4 – The traffic situation is still stable.
- LoS 5 – The lane capacity is full.
- LoS 6 – The section is congested.

Therefore, each location must be categorized individually by the traffic expert.

For this paper, the data describes the specific street in Prague, Legerova Street (3-lane street). The gathered data (speed of traffic flow and traffic flow itself) are stated in Figure 3.

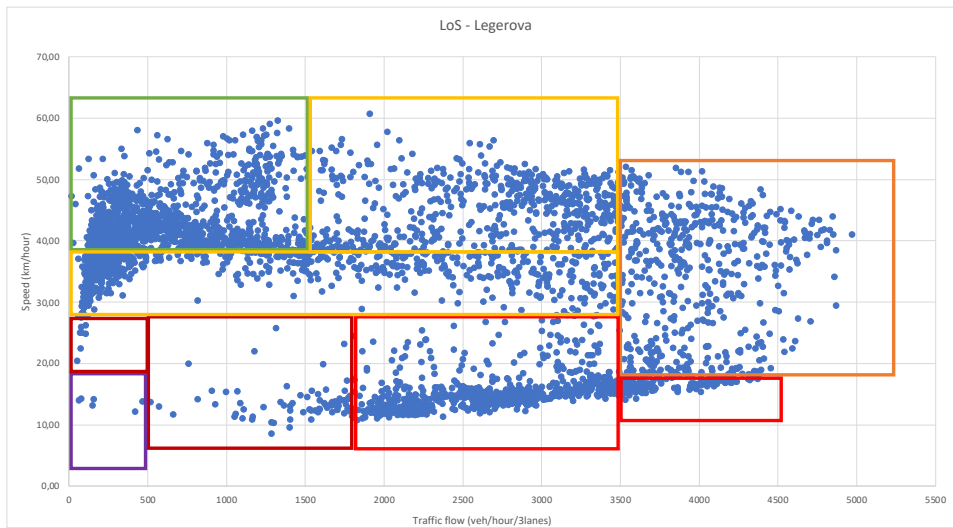

**Fig. 3:** *LoS for Legerova Street in Prague. Green – LoS 1; Yellow – LoS 2; Orange – LoS 3; Red – LoS 4; Burgundy – LoS 5; Purple – LoS 6.*

In this figure, the data are categorized based on the Greenshield fundamental model into 6 categories based on the Czech standards. Based on the expert's evaluation, the areas, shown by rectangles, that belong to the individual LoS are determined. The green rectangle represents the LoS degree equal to 1; sequentially yellow, orange, red, burgundy and purple rectangles represent LoS values 2 to 6.

We take these data and transform them into a 3D graph (see Figure 4).

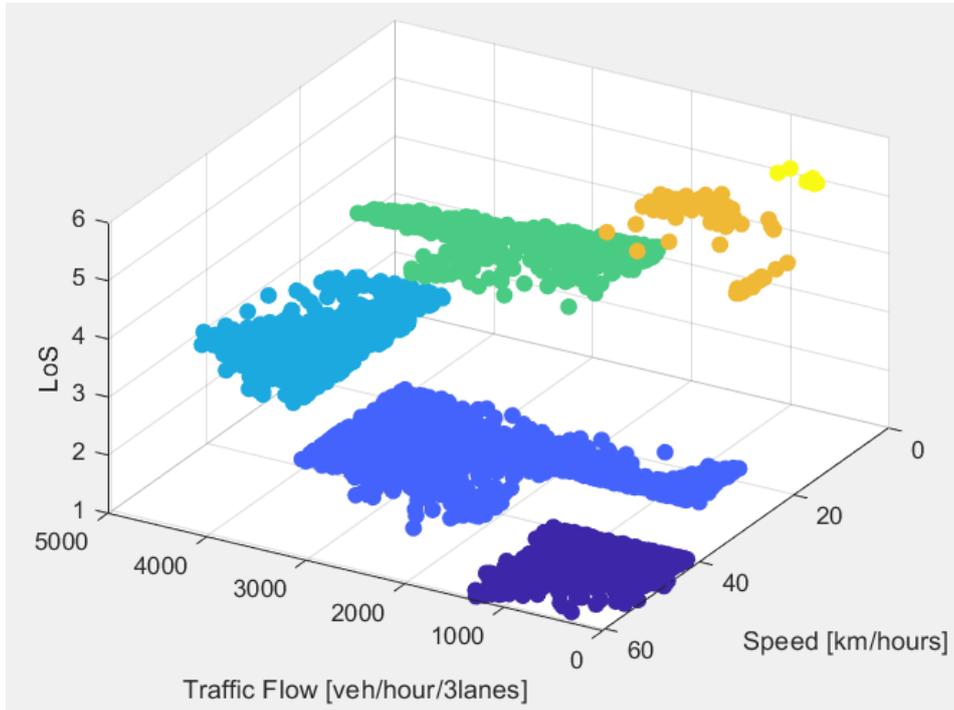

**Fig. 4:** *3D graph of LoS on Legerova Street in Prague.*

In the next step we model the data with the use of the Takagi-Sugeno fuzzy inference system (FIS) [9]. As mentioned before, FIS consists of fuzzy IF-THEN rules. The inputs and the outputs of these rules are specific. Input variables are represented by membership functions which describe fuzzy sets, and output variables consist of polynomial functions. In our study, we used so-called trapezoidal membership functions as input functions and constant functions as output functions. The input "Traffic Flow" is described by six membership functions "Very_Low - Low - Middle - High - Very_High - Extremely_High Flow". The input "Speed" is described by five membership functions "Very_Low - Low - Middle - High - Very_High Speed". As an example, trapezoidal membership functions of fuzzy values for the variable "Speed" are displayed in Figure 5. Constant functions were used as output variables. In the first step of the designing of output variable, it is necessary to determine its domain. We decided to consider the interval [0, 6] and as the values of output, we considered using values one, two, ..., and six. Concerning data displayed in Figure 3, we created 27 rules of type:

"IF Traffic Flow is Very_Low AND Speed is High,
THEN Level_Of_Service is equal to 1".

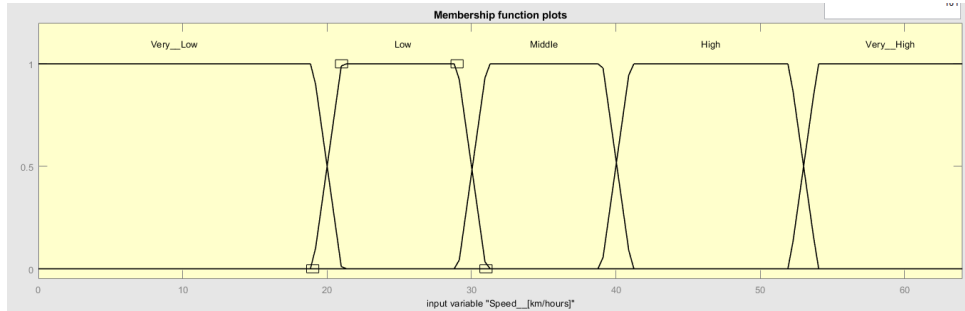

**Fig. 5:** *Example of membership functions of input variable "Speed".*

We process these rules using Takagi-Sugeno FIS. We process them in the MATLAB software, in the Fuzzy Logic Designer toolbox. The output of FIS is the function that represents the approximation of the considered mutual relations between input variables (see Figure 6).

Figure 6 shows that for some input values approximation function reached a value equal to zero. Since the domain of the considered function is a rectangle, some areas are not used as input values in the description of the LoS (see Figure 3). These data values represent anomalies and with the use of these settings for the approximation function, we could detect them very quickly.

Finally, we compare the results obtained by Takagi-Sugeno FIS with the real data. In the first step, we compare data graphically (see Figure 4 and Figure 6). As we can see the designed function has similar progress as the displayed data.

Using the function created by the proposed FIS, we can assign the output "LoS" value to each input pair of "Speed" and "Traffic Flow" values. Therefore, in the second step, we compare the output of the approximation function with the original output of the data pair, value by value. As the output of the approximation function, we mostly get a natural number, while at the borders of two levels of "LoS", we usually get the decimal number. This is also one of the advantages of the proposed system that will be used in the future. The decimal output value means that the considered input pair represents the output that does not strictly belong to one type of LoS. When we want to compare obtained results (the obtained values of LoS) firstly we have to round the obtained decimal numbers to natural numbers. In our example (see Figure 3) we have 3825 input data pairs. After we processed them using an approximation function, rounded them, and compared them with the real output, we got only 28 incorrectly determined values. This represents a 99.27% accuracy of input evaluation.

If we want to process the traffic data generally, we need to know the basic dependencies between data. For example, the number of lanes on one street is very important. This number could be considered as a new input variable in FIS. The information about number of lanes is critical, since such variable defines the capacity of the selected road. The other input data (meteorological, emission data) could be processed in a similar fashion.

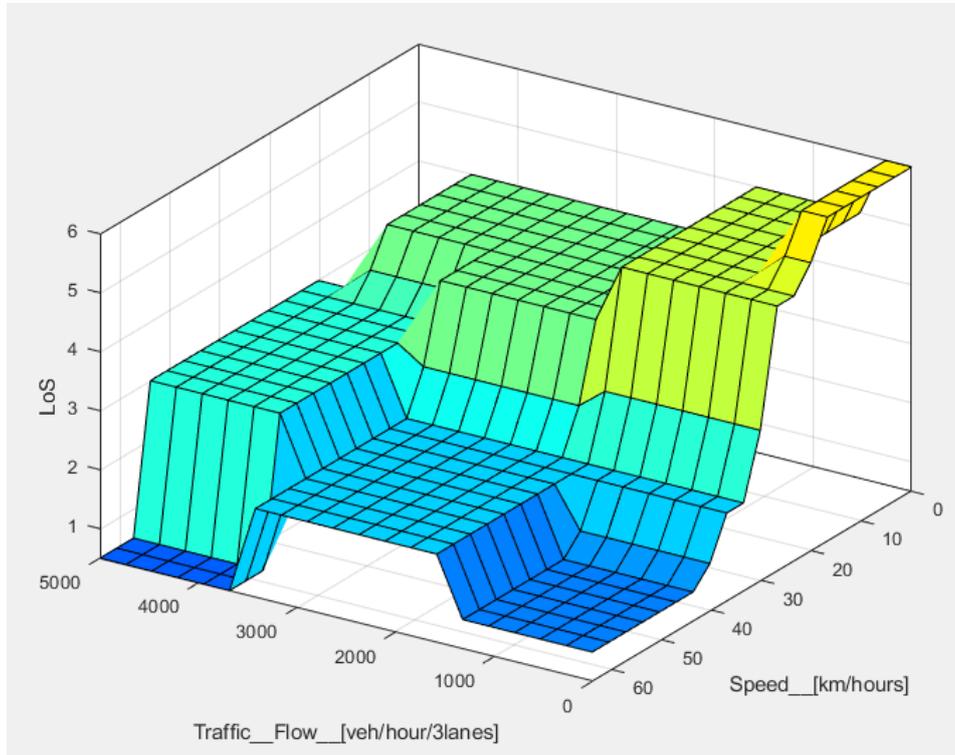

**Fig. 6:** *3D graph of LoS obtained by using Takagi-Sugeno FIS.*

## 4.  Discussion

In our case study, we focused on the relationship between traffic emissions and meteorological conditions using a system approach that integrates various types of data. Findings from previous research point to a significant influence of meteorological conditions on the dispersion and amount of emissions in transport. This influence has fundamental consequences for the planning and management of urban transport, especially in the effort to minimize the impact of transport on the environment.

Presented paper used fuzzy approach in order to model traffic data, which influence Level of Service in the street with the most traffic flow in Prague. These traffic data in combination with meteorological and emission data influence the air quality of the environment and subsequently influence the quality of life and health of citizens in the city. Since the proposed approach reached accuracy of LoS evaluation of 99.27% it can be incorporated into the larger system.

Although the system approach offers a comprehensive view of the issue, it is important to be aware of the potential limitations of the data and models obtained. For example, the variability and unpredictability of meteorological conditions may affect the accuracy of our predictions [10].

# 5.    Conclusion

The results of our study offer valuable insights into the complex relationship between transport and the environment and point to the need for a multidisciplinary approach to solving this issue.

Using fuzzy inference systems, we could create approximation functions, which could help us automatically preprocess the data with high precision. The approximation function could be designed and implemented in such a way that it outputs particular values for anomalous input data. If the output of the function is a natural number, then it means that the input values belong strictly to the considered group of data. On the other hand, if the output of the function is a decimal number, then the input values lie on the boundary between the two considered Levels of Service. This brings important information for the next processing of the data.

Future work related to the presented approach includes:

- use of similar fuzzy model to evaluate meteorological and emission data,
- use of statistical and machine learning models to create predictions based on the fuzzy data,
- estimate actions based on the data analysis outputs.

## Acknowledgement

This work was co-funded by the European Union under the project ROBOPROX (reg. no. CZ.02.01.01/00/22_008/0004590)